\pdfoutput=1

\documentclass[11pt]{article}

\usepackage{acl}

\usepackage{times}
\usepackage{latexsym}

\usepackage[T1]{fontenc}

\usepackage[utf8]{inputenc}

\usepackage{microtype}

\usepackage{multirow}
\usepackage{graphicx}
\usepackage{float}

\title{Benchmarking Azerbaijani Neural Machine Translation}

\author{Chih-Chen Chen \\
  \texttt{chihchen.chen@outlook.com} \\\And
  William Chen \\
  University of Central Florida \\
  \texttt{wchen6255@knights.ucf.edu} \\}

\begin{document}
\maketitle
\begin{abstract}
  Little research has been done on Neural Machine Translation (NMT) for Azerbaijani. In this paper, we benchmark the performance of Azerbaijani-English NMT systems on a range of techniques and datasets. We evaluate which segmentation techniques work best on Azerbaijani translation and benchmark the performance of Azerbaijani NMT models across several domains of text. Our results show that while Unigram segmentation improves NMT performance and Azerbaijani translation models scale better with dataset quality than quantity, cross-domain generalization remains a challenge.
\end{abstract}

\maketitle

\section{Introduction}

With the recent growth in online resources, robust NLP systems have become increasingly available for many of the world's languages. However, this growth has not been enjoyed equally and technologies for many languages are still under-developed, especially relative to the size of their speaker population. This remains the case for morphologically-complex languages, which have been considered a challenge for NLP systems due to the frequency of rare/unknown words. One such example is Azerbaijani, a Turkic language with a highly agglutinative and complex morphology. It has two major varieties: the Northern variant is spoken in the Republic of Azerbaijan, while Southern Azerbaijani regions of Iran. Our experiments focus on Northern Azerbaijani, which is written in Latin script and has considerably more online resources that are able to support the development of NMT systems. 

Little work has been done on NLP systems for Azerbaijani, and even less on machine translation and other generative Seq2Seq tasks. Specifically, there is a lack of benchmarks on the performance of Azerbaijani NMT and the methods that could be used to improve it. Existing studies either include private datasets with unpublished training, testing, and validation splits \cite{maimaiti2022} or solely evaluate on very low-resource scenarios with transfer learning techniques \cite{qi2018}. We build off the approach developed by \citet{guntara-etal-2020-benchmarking}, who sought to develop benchmarks for Indonesian NMT, and extend it to include the evaluation of different pre-processing techniques for Azerbaijani NMT. Our goal is to help address these problems by investigating the following research questions regarding Azerbaijani translation:

\begin{enumerate}
    \item {What segmentation methods work best for Azerbaijani NMT?}
    \item {How important is data cleanliness versus training corpora size for Azerbaijani NMT?}
    \item {How do Azerbaijani translation systems perform across different language domains?}
\end{enumerate}

To answer these questions, we set up the following experiments:

\begin{enumerate}
    \item {We evaluate the performance of different segmentation algorithms to see which perform best for Azerbaijani.}
    \item {We evaluate the effectiveness of scaling to larger training corpora at the cost of alignment quality in Azerbaijani NMT.}
    \item {We categorize open-source Azerbaijani corpora into different domains and evaluate the effectiveness of NMT models trained on individual and multiple domains.}
\end{enumerate}

Our results showed that both the choice of evaluation metric and segmentation algorithm have a large impact in determining which models are the best performing, showing the importance of evaluating across multiple metrics. We also found that sentence alignment quality was a large factor in model performance; the addition of large but noisy/out-of-domain training datasets did not necessarily translate to improved performance. 

\section{Related Work}
Studies on morphologically-complex languages tend to focus on the higher-resource Turkish or extremely low-resource languages like Inuktitut or Quechua. However, there have been many experiments that use Azerbaijani to demonstrate the effects of transfer learning and multilinguality due to its relationship with Turkish. Early MT systems for Azerbaijani were built by \citet{fatullayev2008dilmanc}. Their models were based off of a hybrid between rule-based and statistical machine translation, and could translate to/from English and Turkish. \citet{qi2018} experimented with Azerbaijani in a low-resource setting to improve NMT with aligning pre-trained word embeddings. They showed that including Turkish with Azerbaijani in multilingual NMT significantly improved BLEU score. \citet{neubig2018} explored training paradigms for multilingual NMT that also leverage Turkish to improve Azerbaijani translation. \citet{kim-etal-2019-effective} showed the effectiveness of cross-lingual word-embeddings in improving low-resource Azerbaijani NMT. The most recent work on bilingual Azerbaijani NMT was by \citet{maimaiti2022}, who used Azerbaijani and Uzbek to Chinese translation as case studies for transfer learning with pre-trained lexicon embeddings.

Many studies have been done on the effect on subword segmentation algorithms on downstream NMT. \citet{sennrich-etal-2016-neural} and \citet{kudo-2018-subword} show that such algorithms improve the performance of NMT models using Byte-Pair Encoding (BPE) and Unigram segmentation respectively. While BPE has generally been the standard, recent works show that the Unigram algorithm performs better on agglutinative languages \cite{chen-fazio-2021-morphologically}\cite{richburg-etal-2020-evaluation}\cite{bostrom-durrett-2020-byte}. \citet{mager2022bpe} compared the performance of BPE to morphological segmentation algorithms for indigenous American languages and found that SOTA morphological segmentation methods did not translate to improved performance on NMT. Results in a similar study by \citet{saleva2021effectiveness} were inconclusive when comparing BPE with LMVR \cite{ataman2017linguistically} and MORSEL \cite{lignos2010learning} on Nepali, Sinhala, and Kazakh; the best performing segmentation algorithm was language dependent and the results were statistically indistinguishable. Pre-processing techniques have also been a feature of interest in low-resource translation shared tasks. \citet{chen-fazio-2021-ucf} found that Unigram segmentation \cite{kudo-2018-subword} performed the best for Marathi-English translation at LoResMT 2021 \cite{ojha-etal-2021-findings}. \citet{vazquez-etal-2021-helsinki}  leveraged data cleaning and normalization techniques to overcome differences in orthographic conventions for multilingual models at AmericasNLP 2021 \cite{mager-etal-2021-findings}.

\section{Experimental Setup}

For all of our experiments we use the OpenNMT-py \cite{klein-etal-2017-opennmt} implementation of the Transformer \cite{vaswani2017attention}. We use the set-up from \citet{chen-fazio-2021-morphologically}, which has been shown to perform well with agglutinative languages. The architecture is comprised of 6 encoder/decoder layers, 8 attention heads, size 256 word vectors, and a feed-forward dimension of 2048. The models were trained for 50,000 steps with a batch size of 32.

Translation quality is evaluated using COMET \cite{rei-etal-2020-comet} and the sacreBLEU \cite{sacrebleu} implementations of BLEU \cite{papineni-etal-2002-bleu}  and chrF \cite{popovic-2015-chrf} scores. \citet{kocmi2021to_ship} recommended the use of COMET and chrF, which they found were the metrics that best correspond to human judgement. We also provide BLEU scores due to its standard use in machine translation. Each model was independently trained 10 times such that the presented scores below are the average across all trials.

\subsection{Q1: Segmentation Algorithms for Azerbaijani}
\label{section:q1}

\begin{table*}[htb]
    \centering
    \begin{tabular}{@{}lcccccccc}
        \hline
        \bf \multirow{2}{*}[-.3em]{Segmentation Algorithm} \\
        {} & BLEU & chrF & COMET && \textit{p}-value\\ \hline
        None & 1.596 & 13.136 & -1.205 && \\ 
        BPE & 1.567  & 13.710  & -1.207 && 0.0240\\
        BPE-Guided &  1.517 &  12.010 & -1.234 && 0.0006\\
        PRPE & 1.625 & 13.615 & -1.195 && 0.0099 \\
        Unigram & 1.730 & 14.150 & -1.188 && 0.0013 \\
        \hline
    \end{tabular}
    \caption{\label{seg-score-table}  A comparison of different segmentation algorithms on Northern Azerbaijani to English NMT. Higher scores indicate better performance. \textit{p}-values are calculated using the average COMET score of the given algorithm compared to that of no segmentation.}
\end{table*}
A common pre-processing technique to improve the performance of NLP systems is subword segmentation: separating words into small units to decrease vocabulary size and help the model generalize to unknown vocabulary. The goal of our first set of experiments is to identify which subword segmentation algorithms work best for Azerbaijani. We use the Azerbaijani-English portion of WikiMatrix \cite{schwenk-etal-2021-wikimatrix}, which consists of 276k parallel sentences. The WikiMatrix dataset provides the LASER \cite{artetxe2019massively} score of each sentence pair, which measures the likelihood of a sentence pair being mutual translations. Filtering out sentences with a score less than 1.04 (the recommended LASER threshold) reduces the dataset size to 70,725. The cleaned dataset is then split into 47,385 training sentences, 11,670 validation sentences, and 11,670 test sentences.

Models are trained on text segmented by different techniques: Byte-Pair Encoding (BPE) \cite{sennrich-etal-2016-neural}, BPE-Guided \cite{oretga_nmt_lrl}, Unigram \cite{kudo-2018-subword}, and PRPE \cite{Zuters2018SemiautomaticQW}. BPE and Unigram segmentation are the two most popular segmentation algorithms used in state-of-the-art NMT systems due to their flexibility and ease of use. BPE-Guided \cite{oretga_nmt_lrl} and PRPE \cite{Zuters2018SemiautomaticQW} are morphologically-motivated algorithms that were shown to perform well on NMT for agglutinative languages \cite{oretga_nmt_lrl}\cite{chen-fazio-2021-morphologically}. Prior to subword segmentation, the text is first tokenized by Moses Tokenizer \cite{koehn-etal-2007-moses}.

BPE first splits the corpus into a character level representation. The most frequently occurring pair of tokens are then merged together, a process that is repeated until a pre-defined number of merge operations have been reached. BPE-Guided is an extension of the BPE algorithm that incorporates morphological information through a list of known affixes. BPE-Guided creates a glossary of words that do not contain any known affixes, which is then used by the main BPE algorithm as a list of words to not segment.

Unigram segmentation is a probabilistic segmentation algorithm based on a unigram language model \cite{kudo-2018-subword}. A vocabulary of a pre-defined size is first built by only keeping subwords that least reduce the loss of calculating subword occurrence probabilities via the expectation-maximization algorithm. The output segmentation of a word is then obtained by choosing the most probable segmentation candidate obtained from the Viterbi algorithm \cite{viterbi}.

Prefix-Root-Postfix-Encoding (PRPE) segments a word into three main parts: a prefix, root and a postfix. The algorithm first learns a subword vocabulary of prefixes and postfixes with the help of a language-specific heuristic. PRPE then uses any detected instances of those affixes in a word to extract potential roots and obtain the most probable segmentation of the word.

The BLEU, chrF, and COMET scores are found in Table \ref{seg-score-table}; \textit{p}-values calculated with a paired Student's t-test between a chosen segmentation algorithm's COMET score and the no segmentation baseline are also included. Almost all segmentation methods obtained higher chrF and BLEU scores than the no segmentation baseline. Unigram segmentation performed the best, achieving the highest scores in all three evaluation metrics. PRPE was the second best performing algorithm in BLEU and COMET, but scored lower than BPE in terms of chrF. Interestingly, these two algorithms were also the only ones that performed better than the baseline in terms of COMET score. These results show that both the metric and segmentation algorithm used can have a significant impact on what models are designated as "the best performing", and further encourage the reporting of across multiple evaluation metrics in future work.

\subsection{Q2: Dataset Size vs Cleanliness}
\label{section:q2}

\begin{table*}[htb]
    \centering
    \begin{tabular}{@{}lccccccc}
        \hline
        \bf \multirow{2}{*}[-.3em]{Training Dataset} \\ 
        {} & \# Sentences & BLEU & chrF & COMET\\ \hline
        Clean (T=1.04) & 47,385 & 1.596 & 13.136 & -1.205\\ 
        Slightly Noisy (T=1.03) & 119,725 & 2.276 & 12.614 & -1.292\\ 
        Noisy (T=0) & 252,255 &2.488 & 11.460 & -1.399\\
        \hline
    \end{tabular}
    \caption{\label{tradeoffTable}  A comparison of the tradeoff between dataset size and cleanliness. T is the LASER score threshold use to filter sentence pairs, which is a measurement of the likelihood that two sentences are mutual translations.}
\end{table*}
We conducted a second set of experiments to examine the tradeoff between dataset cleanliness and dataset size in regards to NMT performance by using the alignment scores provided by the WikiMatrix dataset \cite{artetxe2019massively} as a measurement of cleanliness. To do so, we created additional training datasets with the WikiMatrix sentence pairs left unused in Section \ref{section:q1}. We combine these remaining sentences with the clean 47k sentence training set to form a noisy 252k sentence training dataset. As a middle ground, we also create a third training dataset of 120k sentences by only keeping sentence pairs with a score of at least 1.03 from the large noisy dataset. The validation and test sets are reused from \ref{section:q1}. The text was not pre-processed with any subword segmentation algorithm to isolate any impact on the performance metrics to the change in training data.

The results (Table \ref{tradeoffTable}) provide an interesting reflection of how the evaluation metrics are calculated. BLEU \cite{papineni-etal-2002-bleu} scores increased as the training dataset size grew, but chrF \cite{popovic-2015-chrf} and COMET \cite{rei-etal-2020-comet} scores decreased. We hypothesize that this is because the additional training data increased the vocabulary size of the model and thus allowed it to recognize otherwise unknown words in the test set. Our results corroborate the findings of \citet{kocmi2021to_ship} and show the inaccuracy of BLEU compared to other metrics: evaluating only with BLEU would indicate that training on the smaller dataset was worse despite the opposite holding true.

\subsection{Q3: Domain Benchmarks}

Our final experiment was to evaluate the performance of an Azerbaijani NMT model across several domains of text. We first obtained all Azerbaijani-English (az-en) data from OPUS \cite{tiedemann-nygaard-2004-opus}, which consist of the following parallel corpora: WikiMatrix \cite{schwenk-etal-2021-wikimatrix}, CCMatrix \cite{schwenk-etal-2021-ccmatrix}, Tatoeba, ELRC public corpora, Tanzil, GNOME \cite{TIEDEMANN12.463}, QED \cite{abdelali-etal-2014-amara}, TED2020 \cite{reimers-2020-multilingual-sentence-bert}, and XLEnt \cite{elkishky_xlent_2021}. The corpora were categorized by domain, of which the domains with little data (lecture, news, and tech) were aggregated into a larger ``Mixed" domain dataset. We thus evaluate the model on four different datasets: General (1,325,660 lines), Religious (269,445 lines), Entities (298,236 lines), and Mixed (68,256 lines). Each dataset was then split into 66.7\% training sentences, 16.6\% validation sentences, and 16.6\% test sentences. All text is pre-processed with Moses Tokenizer \cite{koehn-etal-2007-moses} and segmented with a Unigram segmentation model \cite{kudo-2018-subword}.

\begin{table}[htb]
    \begin{center}
    \begin{tabular}{|l|r|r|}\hline
        Dataset   		                & \# Sentences & Domain    \\\hline
        CCMatrix                        & 1,251,255     & General \\
        WikiMatrix (T=1.04)   	        & 70,725    	& General \\
        Tatoeba                         & 3,680         & General \\
        \hline
        ELRC                            & 129           & News \\
        \hline
        Tanzil                          & 269,445       & Religious \\
        \hline
        GNOME                           & 40,075        & Tech \\
        \hline
        QED                             & 16,442        & Lecture \\
        TED2020                         & 11,610        & Lecture \\
        \hline
        XLEnt                           & 298,236        & Entities \\
        \hline
        \textbf{Total}                  &\textbf{1,961,597
        }&  \\\hline
    \end{tabular}
    \end{center}
    \caption{Dataset Statistics}
    \label{dataset}
\end{table}

We independently train models on each dataset. To evaluate the system's ability to generalize across domains, we train another model on the data combined across all 4 datasets. The models are trained for 300,000 steps and are evaluated using the best performing checkpoint on the validation set. The 4 domain-specific models are evaluated on the test set of their domain and the model trained on combined data is evaluated on each domain.

\begin{table*}[htb]
    \centering
    \begin{tabular}{@{}lccccccc}
        \hline
        \bf \multirow{2}{*}[-.3em]{Test Set} & \multicolumn{3}{c}{Trained on Domain Only} & \multicolumn{3}{c}{Trained on Combined Data} \\
        {} & BLEU & chrF & COMET & BLEU & chrF & COMET \\ \hline
        General & 5.55  &  16.999 &  -1.069 &  3.981 &  14.795 &  -1.1658 \\ 
        Religious & 23.199  & 44.535 & -0.818 & 17.285  & 34.285 &  -0.6010 \\
        Entities & 7.607 & 19.845 & -0.929 &   1.279 & 11.428 &  -1.1751 \\
        Mixed &  22.725 &  35.648 & -0.136  & 4.555 & 15.293 & -1.0216 \\
        \hline
    \end{tabular}
    \caption{\label{domain-score-table} A comparison of the BLEU, chrF, and COMET scores between models trained on a specific data domain and a model trained on data across all domains.}
\end{table*}

Most of the domain-specific models performed better than the model trained on combined data (Table \ref{domain-score-table}). An exception was on the Religious dataset; while the Religious model performed better than the Combined Data model in terms of BLEU and chrF, the Combined Data model achieved a better COMET score. This indicates that training on a more general dataset allowed the model to output more words that were closer to the label translation in the embedding space (higher COMET score) but differed in terms of subwords/characters used (lower BLEU and chrF score). These results also corroborate those of \ref{section:q2}, again showing the importance of data cleanliness. Models trained on the smaller and cleaner Religious and Mixed datasets performed better than those trained on the larger General, Entities, and Combined datasets. The result is particularly noticeable with the Mixed dataset model, which achieved a COMET score of -0.136 despite having only 45,500 training sentences.

\section{Conclusion}
We trained several Azerbaijani NMT models on text segmented by different algorithms and show that using Unigram segmentation can noticeably improve translation quality. We also demonstrate that properly cleaning data can lead to significant gains in performance, even when shrinking the training corpora. Finally, we evaluated the performance of Azerbaijani-English NMT models across multiple domains. Our results demonstrate that while generalizing across domains remains a challenge for Azerbaijani NMT, specialized models are still able to achieve a competitive performance. 

\section{Future Work}
Our experiments focused only on Northern Azerbaijani due to scarcity of data for the Southern variant. One route for exploration to develop NMT systems for the latter is to compare the effectiveness of lower-resource cross-dialectal transfer from Northern Azerbaijani against higher-resource cross-lingual transfer from Turkish. Developing NMT systems for Southern Azerbaijani is particularly challenging since it is written in Arabic script, introducing the need for transliteration to properly take advantage of transfer learning paradigms. Further evaluation could also be done on the transfer learning and multilingual techniques used to improve Azerbaijani translation introduced in previous works. While those studies show that such techniques are able to improve translation quality over a simple baseline, there are little to no comparisons of their effectiveness relative to each other.

\bibliography{acl_latex}
\bibliographystyle{acl_natbib}

\end{document}